# Point-supervised Single-cell Segmentation via Collaborative Knowledge Sharing

Ji Yu

**Abstract**—**Despite their superior performance, deep-learning methods often suffer from the disadvantage of needing large-scale well-annotated training data. In response, recent literature has seen a proliferation of efforts aimed at reducing the annotation burden. This paper focuses on a weakly-supervised training setting for single-cell segmentation models, where the only available training label is the rough locations of individual cells. The specific problem is of practical interest due to the widely available nuclei counter-stain data in biomedical literature, from which the cell locations can be derived programmatically. Of more general interest is a proposed self-learning method called collaborative knowledge sharing, which is related to but distinct from the more well-known consistency learning methods. This strategy achieves self-learning by sharing knowledge between a principal model and a very light-weight collaborator model. Importantly, the two models are entirely different in their architectures, capacities, and model outputs: In our case, the principal model approaches the segmentation problem from an object-detection perspective, whereas the collaborator model a semantic segmentation perspective. We assessed the effectiveness of this strategy by conducting experiments on LIVECell, a large single-cell segmentation dataset of bright-field images, and on A431 dataset, a fluorescence image dataset in which the location labels are generated automatically from nuclei counter-stain data. Implementing code is available at https://github.com/jiyuuchc/lacss**

## I. INTRODUCTION

AUTOMATED cell segmentation from microscopy images is the first step in the pipeline of single-cell analysis. In recent years, segmentation methods based on deep learning have demonstrated unparalleled performance and are increasingly being adopted by biologists as the method of choice [1]–[3]. Currently, there are two general types of single-cell segmentation models in the literature. The first type treats the problem as a pixel-level classification/regression task [2]–[9], which has its roots in the semitic segmentation field. In the simplest case, the model classifies each pixel of the input microscopy image as either the foreground, the background, or the cell border. A simple post-processing step can then be employed to create the segmentation masks of individual cells. Unfortunately, cell border classification often suffers from high error rates. More recent models in this category typically perform more sophisticated pixel mappings, e.g., to the

Euclidean distance to the nearest background pixel. Nevertheless, the general idea is to convert a microscopy image to an intermediate pseudo-image that is more algorithmically manageable, and to use hand-crafted post-processing algorithm to convert the pseudo-image into instance segmentations of single cells. The second general approach is to use an object detection/segmentation model, e.g. MaskRCNN [10]. These models have an object detection branch that is trained to predict the bounding-boxes of the objects (i.e., cells) in the image, as well as a relatively light-weight segmentation branch to produce the exact segmentation mask within each bounding-box. Even though these models were not designed specifically with biomedical data in mind, their applications to biomedical imaging data [1], [11] appears to be straightforward.

While both types of models perform well for the single-cell segmentation task, they also suffer from the disadvantage of needing large-scale training data, which can be very expensive to produce. In response, recent studies have focused on two revenues in attempts to mitigate this problem. The first line of research involves domain adaptation [12]–[17]. The goal is to adapt a target domain dataset to a labeled source domain dataset. This allows for producing new models on dataset that is unlabeled or mostly unlabeled, assuming a well-annotated source dataset is available. A second line of research aims to train models with weak supervisions [11], [18]–[21], using approximate labels instead of full segmentation masks, which would significantly reduce labeling cost.

This paper focuses on a specific weakly-supervised training setting where the only available label is the rough locations of individual cells. This specific setting is of practical relevance, particularly in the fluorescence imaging domain, due to the common practice of acquiring nuclei counter-stain image when collecting cellular microscopy data, in which case the point labels can usually be computed from the nuclei image directly without human input. Many authors have proposed methods in attempts to utilize point labels [18], [22]–[25], but a robust and generalizable method for training single-cell segmentation models has yet emerged. Several of the published methods [22]–[24] rely on the assumption that there is color consistency within the instance, which is rarely true for cellular imaging data; and methods [18] that doesn't assume color consistency produce image-level segmentations instead of instance-level single-cell segmentations.

Here we propose a new method to train a point-supervised instance segmentation model, $\mathcal{X}_P$, based on a general architecture outlined in Fig. 1:

Ji Yu is with University of Connecticut, School of Medicine, Farmington, CT 06030 (e-mail: jyu@uchc.edu)
.

$$\mathcal{X}_P : \boldsymbol{x}; \boldsymbol{\theta_p} \mapsto \{\boldsymbol{r_i}, \boldsymbol{m_i} \mid i = 1 \dots n\} \qquad (1)$$

where $\boldsymbol{\theta_p}$ is the model weights, $\boldsymbol{x} \in \mathbb{R}^{H \times W \times C}$ represents the input image, $\boldsymbol{r_i} = \{y_i, x_i\}$ is the prediction of a single cell's location by the detection head (Fig. 1), and $\boldsymbol{m_i} \in [0, 1]^{H \times W}$ is the probability map of the cell's segmentation mask, outputted by the segmentation head of the model (Fig. 1).

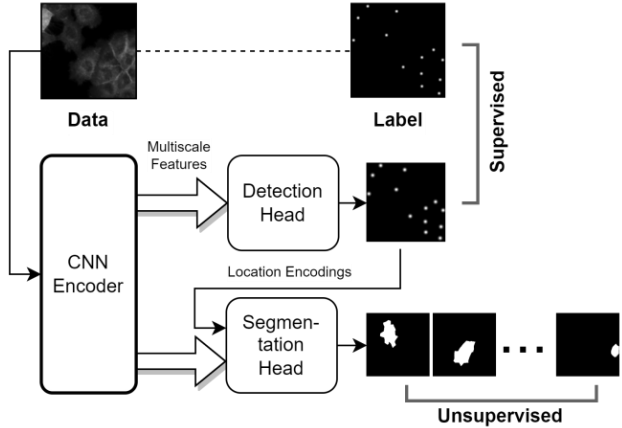

Fig. 1. Outline of the principal model architecture.

Since training the detector branch of the model is straightforward with point supervision, the main challenge is to find a way to train the segmentation branch in a self-supervised manner. One of the contributions of this paper is to demonstrate a novel self-supervised training technique we call collaborative knowledge sharing (CKS). In this training scheme, we utilize a pair of what we call principal/collaborator models. The principal model is defined as (1), which can be viewed as a subtype of object detection model. The collaborator model, on the other hand, performs the instance segmentation task from a pixel-level classification perspective:

$$\mathcal{X}_C : \boldsymbol{x}; \boldsymbol{\theta_c} \mapsto \{\boldsymbol{M_c}, \boldsymbol{B_c}\} \qquad (2)$$

where $\boldsymbol{M_c} \in [0, 1]^{H \times W}$ is the prediction of the binary segmentation for the whole input image, i.e., for all cells combined, and $\boldsymbol{B_c} \in [0, 1]^{H \times W}$ represents the predictions of cell-border pixels. In addition, the collaborator model is of significantly lower capacity than the principal model. Self-learning is achieved by training the two models to produce consistent outputs from the same input (Fig. 2).

As discussed in the beginning of this section, both the object-detection type and pixel-mapping type models can be effective in performing the single-cell segmentation task. However, it is not known previously that they can be combined in this way to form an effective strategy for self-learning. This training scheme can be viewed as a subtype of the consistency regularization method [26], which has gained popularity recently in both semi-supervised [27] and unsupervised [28] training settings. The goal of consistency regularization is to minimize the differences between outputs of multiple computational paths despite added noises. The noise can be at the sample level by data augmentation, or at the computational path level via stochastic operations, e.g., dropout. However, different paths typically have similar structure and computational complexity. The key distinction feature of CKS is that the principal model and the collaborator model have entirely different architectures. In addition, the collaborator model is not intended to become "competent" at its task. Instead, we adopted the principal-collaborator metaphor to highlight that the key contribution of the collaborator model is to bring in a different perspective to the problem at hand, resembling a human collaborator in the academic setting. In addition, here we also report characteristics of CKS that are unexpected of those of traditional consistency regularization. For example, we found that it is preferable to keep the collaborator model at a significantly lower capacity in comparison to the principal model, which suggested that there are divergent mechanisms at play between CKS and consistency regularization.

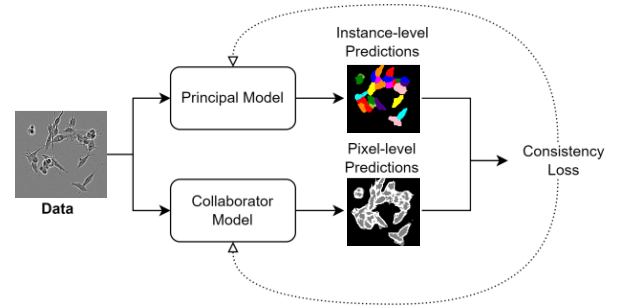

Fig. 2. Diagram illustrating the concept of collaborative knowledge sharing training.

## II. Related work

### A. Weakly-supervised instance segmentation

Weakly-supervised learning reduces the annotation cost by employing incomplete or noisy labels to train models. In the segmentation literature, many weak labels have been proposed in place of segmentation masks, e.g., scribbles [29], [30], bounding boxes [20], [31], and points [23], [32]. For instance-level segmentations, bounding box in particular fit neatly into the training pipeline of the object detection/segmentation models. In addition, [31] shows that bounding box can be viewed as a noisy segmentation mask, and recursive training, i.e., using earlier training results as the updated labels for later training, is an effective denoise strategy. A similar strategy was employed in [18] with point supervision. The authors obtained good results with image segmentation after combining recursive training with co-training of model pairs, although the method was not generalized to instance segmentation.

An alternative line of research focuses on performing training on simple auxiliary tasks [25], [33]–[35], such as image classification, and try to obtain segmentation results by examine the model structure. For example, the class activation mapping (CAM) [36] is a popular proxy for segmentation masks. This strategy allows very simple labeling, e.g., image-level labeling. However, the segmentation results are usually approximate.

### B. Point supervision.

In the segmentation literature, the term point supervision has been used to refer to two distinct types of labels. The first type requires randomness in location and should be in fact viewed as an extremely under-sampled segmentation mask [19], [32]. For this reason, this type of point label is usually directly used as a replacement for segmentation masks without additional changes in the model or the training pipeline. In addition, these works employed additional bounding-box labels to learn object detection in order to achieve instance segmentation.

Most relevant to our work are models trained only on non-randomized point labels. For example, [22], [23] both proposed methods to train nuclei segmentation models by deriving pseudo-segmentation label from point labels. However, the methods rely on color consistency within the nucleus, making it difficult apply them to other cell segmentation problems. In [18] the authors proposed a method based on consistency regularization and recursive training. But the model produced image-level segmentation instead of single-cell segmentation. [25] trained a model using only point label and used CAM to approximate cell segmentation. We ourselves have published a method [21] to train a single-cell segmentation model by combining point labels with an image-level segmentation label, which is a direct precursor to the work discussed here.

Finally, point labels had also been used for delineating purpose during inference [37].

### C. Consistency regularization and collaborative learning

Our CKS training method should be viewed as a subtype of the consistency regularization method [26], [27]. These methods use consistency loss, which can be defined on both labeled and unlabeled inputs, to regularize model training. A typical setup involves the co-training of a pair of teacher/student models with the same architecture. CKS deviates from this typical approach by using two models of intentionally different architectures. In addition, it is more natural to view the entire collaborator model as a deep-regularization term, which controls the search space of the principal model. Among the many variants of consistency regularization methods, the one we resemble the most is probably cross-task consistency learning [38], which has found various applications in biomedical segmentation [39]–[41]. This method trains a multi-task model with both shared and divergent computational paths and is based on inference path invariance. The underlying logic is that the predictions made for different tasks out of one input image should be consistent. Like CKS, cross-task consistency compares divergent output types. Different from CKS, different computational paths in cross-task consistency learning are still relatively balanced and usually share weights.

The concept of knowledge sharing has its roots in knowledge distillation [42], which was originally designed to transfer knowledge from a large teacher model to a smaller student mode. The more recent extension of this concept to online model training [43] is a form of collaborative learning, for which there is no clear distinction between the teacher and the student. In that sense, CKS is closer to the original form of knowledge distillation, although to arrive at an entirely different purpose.

## III. METHOD

### A. Principal model

The architecture of our principal model's design (Fig. 1) resembles that of an object detection/segmentation model, except the detection branch produces predictions of cell locations instead of bounding boxes. We chose ConvNeXt [44] as the CNN encoder backbone due to its good performance in various machine vision tasks and use feature pyramid network (FPN) to integrate ConvNeXt output, forming the multi-scale image features as inputs for the two branches of the decoders. Different from standard instance segmentation models, our model does not expand range of feature scales at the FPN stage. We reasoned that this is not necessary because the typical instance sizes in a single-cell segmentation problem would not be very large.

The detection head is a standard multi-layer CNN, which outputs $D \in [0,1]^{H \times W/s/s}$ representing the probability of finding a cell at any of the grid $(\frac{H}{s} \times \frac{W}{s})$ locations, as well as an $D_s \in \mathbb{R}^{H \times W/s/s}$, representing the relative offset of an exact cell location within each grid. Here the $s$ is the scaling factor of the feature input relative to the source image, and is either 4, 8, 16 or 32 in our model setup. The model performs detections on all scales of the multi-scale feature map, although the detectors share weights at different scales. Note that unlike the standard bounding-box-based object detection scheme, in which different scales are used for the detections of instances of different sizes, here we detect all instances at all scales, simply because we do not have any labeling information regarding the instance sizes. At the inference time, redundant detections are removed by non-max-suppression, using $D$ for ranking.

Training the detector branch is straightforward, since the ground truth values, $D^{GT}$ and $D_s^{GT}$, can be computed from the point labels. We use focal binary cross-entropy loss for $D$ and $L2$ loss for $D_s$, i.e.:

$$\mathcal{L}_{det} = \sum_s \mathcal{L}_{fce}(D, D^{GT}) + \|D_s - D_s^{GT}\| \qquad (3)$$

where $\mathcal{L}_{fce}(\cdot)$ is the focal binary cross-entropy loss function. We also perform a minor label smoothing by considering grid locations within a small threshold distance to the true location as positive, even if the cell location is not exactly within the grid.

The segmentation head computes $\{m_i \mid i = 1 \dots n\}$ directly, using the image features at the highest resolution (s=4) as the input. Here $n$ is the number of cells. Therefore, the head performs $n$ parallel segmentation computations for each input image. For the datasets we worked with in this paper, the $n$ value can be as large as 3000-4000. Therefore, computational efficiency is important here. Since we do not know the instance sizes, we cannot use algorithms such as Roi-Align to define the regions of segmentation analyses. On the other hand, performing segmentation on the whole image is costly and unnecessary. Instead, we use a model hyperparameter to define the maximum area surrounding the cell location for segmentation computation. The $\boldsymbol{m_i}$ values outside the region are assumed to be 0. Because the analyzed area almost always encompasses more than one cell, we need to incorporate the cell location information into the feature inputs to break the translational invariant property of CNN, and thus output the segmentation for a specific cell. The model does this by creating a position encoding tensor derived from the feature vector at the location of the cell, feeding it through a multi-layer perceptron (MLP) and reshaping and resizing the resulting vector to match the size of the segmentation window. The position encoding tensor is concatenated to the image feature tensor to form the full input for segmentation prediction (Fig. 3).

It is easy to see that if the training data were labeled with segmentation masks, the principal model can be easily trained in a fully supervised manner by employing cross-entropy loss to the segmentation output.

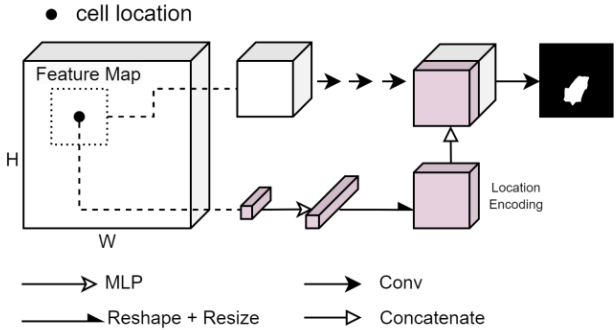

Fig. 3 Design of the segmentation head of the principal model. The inputs are the image feature map (from CNN encoder) and a cell location (from the ground truth label during training and the detection head during inference). The location encoding is computed from the feature vector at the cell location and was concatenated to the feature map to define the specific instance that needs to be segmented. MLP: multi-layer perceptron. Conv: a 2D convolution layer with activation.

### B. Collaborator model

Our collaborator model is a very light-weight CNN performing the pixel level mappings. To facilitate experimentations, here we use two separate nets to compute $\boldsymbol{M_c}$ and $\boldsymbol{B_c}$ with no shared weighs, although this is not necessary in real applications. We use a U-Net-like CNN to compute $\boldsymbol{M_c}$, following other works that employed models of this type, although our default net is much shallower (two down-scale operations total) and narrower (16, 32, and 64 feature channels at three respective scales). The default net for computing $\boldsymbol{B_c}$ is even simpler: it is a three-layer CNN with 32 channels each. The design choices reflect our belief that the contribution of the collaborator model is not to be *good* at the segmentation task, but to offer a different perspective to the problem.

In addition, the collaborator model was discarded after training and not used during inferences.

### C. Knowledge Sharing - Segmentation

To entice the collaborator model to learn segmentation from the principal model, we first need to construct an image-level segmentation prediction from the instance level outputs. We do this by finding at each pixel the maximum of instance segmentations logit, at the same time shifting the prediction by an offset representing the prior knowledge:

$$\boldsymbol{M_p} = \max_i \sigma[\ell(\boldsymbol{m_i}) + \pi \ell_{prior}(d, y_i, x_i)] \qquad (4)$$

Here the $\sigma(\cdot)$ is the sigmoid function, $\ell(\boldsymbol{m_i})$ represent logits of $\boldsymbol{m_i}$, and $\ell_{prior}$ is a 2D Gaussian shape centered at $y_i, x_i$ with a variance $d^2$, and $\pi$ a small scaling factor reflecting the confidence of the prior. The $\ell_{prior}$ term allows us to incorporate the point label information into the trainings of the collaborate model. Additionally, the hyperparameter $d$ allows one to express a prior knowledge regarding the expected cell size, which we found to be helpful in achieving better model accuracy (see ablation study).

We train the collaborator model to follow $\boldsymbol{M_p}$. While the first instinct is to use a cross-entropy loss for this, e.g.,:

$$\mathcal{L}_{CE} = \frac{-1}{H \times W}\Big[\left(1 - \mathbf{1}_{M_p \geq 0.5}\right) \cdot \log(1 - \boldsymbol{M_c}) + \mathbf{1}_{M_p \geq 0.5} \\ \cdot \log(\boldsymbol{M_c})\Big] \qquad (5)$$

However, a disadvantage is that one needs to convert the continuous soft-label $\boldsymbol{M_p}$ to a discrete hard-label (i.e., 0 or 1) for this approach. We found that (see results of ablation studies) the training tends to be trapped in a suboptimal state in this case, and a better approach is to use a simple linear loss function defined as:

$$\mathcal{L}_M = \frac{1}{H \times W}\Big[(1 - \boldsymbol{M_p}) \cdot \boldsymbol{M_c} + \boldsymbol{M_p} \cdot (1 - \boldsymbol{M_c})\Big] \qquad (6)$$

While not a commonly used loss function, its effect can be intuitively seen by examine its gradient:

$$\frac{\partial \mathcal{L}_M}{\partial \boldsymbol{M_c}} = \frac{1 - 2\boldsymbol{M_p}}{H \times W} \qquad (7)$$

Thus $\boldsymbol{M_c}$ will move towards either 0 or 1 depending on whether $\boldsymbol{M_p}$ is bigger or smaller than 0.5. In other words, its general effect is almost identical to that of a cross-entropy loss function, but unlike the cross-entropy loss, the gradient is higher where $\boldsymbol{M_p}$ values are closer to either 0 or 1, and lower where $\boldsymbol{M_p} \approx 0.5$. In other words, the training primarily focuses on the "confident" $\boldsymbol{M_p}$ predictions and ignores the "ambiguous" predictions.

Conversely, to transfer knowledge from the collaborator to the principal model, we use:

$$\mathcal{L}_S = \frac{1}{K}\left\{\sum_i (1 - M_c) \cdot m_i + M_c \cdot (1 - m_i) \right.$$
$$\left. + \sum_i m_i \cdot \sum_{j \neq i} m_j \right\} \tag{8}$$

The first part of $\mathcal{L}_S$ resembles (6), but is computed at instance level, and has the effect of driving $m_i$ toward $M_c$. This term alone, however, will lead to over-segmentation, because $M_c$ is the segmentation of all cells. To compensate, the second half of $\mathcal{L}_S$ incurs a penalty whenever two instances assign non-zeros values at the same pixel. $K$ is a normalization factor.

### D. Knowledge Sharing – Cell border

Similar as the last section, we first need to reconstruct an image-level prediction from the instance level segmentations:

$$B_p = \tanh \sum_i \varphi(m_i) \tag{9}$$

where $\varphi(\cdot)$ is the sobel filter, which convert the segmentation foreground to segmentation edges, and the hyperbolic tangent function (tanh) was applied to ensure that the results remain bound between 0 and 1. We used L2 loss to train both the collaborator model and the principal model:

$$\mathcal{L}_B = \|B_c - B_P\| \tag{10}$$

### Preventing Model Collapse

Model collapsing is a common issue in self-supervised learning schemes, which we need to prevent. The most likely collapsing mode in our setting is that all segmentation predictions producing zero everywhere. We incorporate a simple penalty term to prevent this happening:

$$\mathcal{L}_{MC} = \frac{\delta}{n}\sum_i (\overline{m_i})^{-1} \tag{6}$$

where $\overline{m_i}$ denotes that average of $m_i$, and $\delta$ is small scaling factor to ensure that the term is usually much smaller than other losses, except when the predictions for all pixels move towards zeros.

### E. Model Updating

Both the principal model and the collaborator model are updated using the standard gradient descent method every batch. But the two models learn with different loss functions. For the principal model:

$$\mathcal{L}_{pr} = \lambda_{det}\mathcal{L}_{det} + \lambda_S\mathcal{L}_S + \lambda_B\mathcal{L}_B + \lambda_{MC}\mathcal{L}_{MC} \tag{7}$$

And for the collaborator model:

$$\mathcal{L}_{co} = \lambda_M\mathcal{L}_M + \lambda_B\mathcal{L}_B \tag{8}$$

Here various $\lambda$ values are relative loss weights. In this study we do not tune these values and set all $\lambda$ weights to be 1.

## IV. Experiments

### A. Datasets

We conducted experiments on LIVECell [1] and A431 [45] datasets.

The LIVECell dataset is a large-scale microscopy image dataset aimed at testing single-cell segmentation models and is currently the largest dataset available of this type. The dataset consists of bright field images of $520 \times 704$ from eight different cell lines exhibiting variable cell morphology and imaging contrasts. The data were also collected on samples with a very wide range of cell density. The original authors had split the dataset into training (3253 images), validation (570 images), and testing (1564 images) splits.

The A431 dataset was a fluorescence microscopy dataset. The main interest in this dataset is that it contains point labels that had been generated in an unsupervised fashion from the nuclei counter-stain data. Therefore, the entirely training pipeline on this dataset can be considered a fully *unsupervised* one. The training set contains 500 images of human squamous-cell carcinoma A431 cell of $512 \times 512$ pixels. In addition to the point label, the training set was also labeled with manually produced image-level segmentations, which we did not use. The test contains 25 images that had been manually segmented.

### B. Implementation Details

#### 1) Networks and training configurations

All experiments employed a ConvNeXt backbone at the small configuration with a patch size of 4. We set the maximum segmentation area to be relatively small ($96 \times 96$ pixels) to allow quick experimentation.

For the LIVECell dataset, we trained models under both the weakly-supervised setting (using point labels) as well as the fully-supervised setting (using segmentation mask). The supervised model presumably sets the upper bound of model performances. To assign the hyperparameter $d$ in (4), we grouped the eight cell lines into three groups according to their average size (large, medium, and small) and used three different values (20, 15 and 10) respectively. The value of $\pi$ is fixed at 2.0 for all experiments. We train $9 \times 10^4$ steps with one image per step, using ADAM optimization with an initial learning rate of $10^{-3}$ and a finetune rate of $2 \times 10^{-4}$. We use two different model initialization schemes for the semi-supervised model: a) We randomly initialized the model using the He method [46], b) We pretrained the principal model on TissueNet dataset [2], which is a fluorescence microscopy dataset on tissue cryo-sections. We will use CKS-1 and CKS-2 to denote models trained under these two different initialization schemes, respectively.

For the A431 dataset, we performed similar experiments except no fully supervised experiments. Since data is from one cell line, we set $d$ to be a constant 20. All other training settings are the same as LIVECell.

### 2) Preprocessing and augmentation

The LIVECell dataset was labeled with instance segmentations. We first computed centroids of each segmentation and used these values as the point labels. The dataset included cell lines of very different sizes. We pre-scaled all images so that the largest cell lines (SKOV3 and Huh7) were scaled down 30% and the smallest BV2 and MCF7 lines up by two folds. This allows us to set the maximum area for segmentation to a smaller value (96 pixels), which speeds up the experiments. For both the LIVECell and A431 datasets, we use a simple augmentation protocol that includes image rotation, flipping and resizing (±15%). Additionally, the augmentations also include random adjustments of brightness and contrast.

### 3) Evaluation

We evaluate the model performance using the Average Precision ($AP$) metric:

$$AP_{IOU} = \frac{\sum_k T_k P_k}{C} \tag{9}$$

where $C$ is the total number of ground truth cells, $T_k$ is an indicator function of whether the k-th detection is positive or negative according to the specified segmentation mask $IOU$ (intersection-over-union) threshold, and $P_k$ is the precision of the first k detections. We do not perform smoothing of the precision-recall curve when computing $AP$. We also computed mAP, which is the average of APs at the series of ten $IOU$ thresholds ranging from 0.5 to 0.95 at an equal spacing.

Additionally, we also compute instance $Dice$ metric, defined as:

$$Dice = \frac{1}{2}\left(\sum_i^{N_g} w_i D(\boldsymbol{g}_i, \boldsymbol{m}_{gi}) + \sum_j^{N_p} w_j D\left(\boldsymbol{m}_j, \boldsymbol{g}_{m_j}\right)\right) \tag{15}$$

where $\boldsymbol{g}_i$ is the $i$-th ground-truth cell, $\boldsymbol{m}_j$ is the $j$-th model prediction, $\boldsymbol{m}_{gi}$ and $\boldsymbol{g}_{m_j}$ are the matched segmentations of the prediction and the ground-truth object, $D(\cdot)$ the dice operator, $N_g$ and $N_p$ the number of ground-truth and predicted objects, and $w_i$ and $w_j$ area-adjusted weighting coefficient. The $Dice$ metric is useful for comparing with models that do not produce ranked outputs, for which the $AP$ metrics are undefined.

### 4) Baselines

We compare with several state-of-the-art baseline methods. For fully-supervised model training, we compare with cellpose [3], a popular CNN model representing the state-of-the-art using the pixel-mapping strategy for single-cell segmentation, and MaskRCNN [1], representing the state-of-the-art for object-detection strategy. In both cases, we use published models trained on LIVECell dataset, and compute additional metrics when necessary.

For point-supervised training on LIVECell dataset, we compare with method described in [25] (Nishimura-1) and in [47] (Nishimura-2). The former is a CNN model trained with only point-labels and employ a CAM-based postprocessing step to extract cell segmentations. The latter combines point label with image-level classification label (cell types), and employed a unique instance paste data augmentation scheme to improve the result. Evaluation of both methods on LIVECell dataset had already previously published [47].

For point-supervised training on A431 dataset, we compared with Nishimura-1 method by applying the publicly available implementation to the A431 dataset. In addition, because this training pipeline does not employ manual labeling, we also compare with PDAM, an unsupervised domain adaptation (UDA) method described in [14], using the TissueNet dataset as the source domain data for model training.

### C. Fully Supervised LIVECell Models

We first train a fully supervised principal model on LIVECell dataset to verify that its overall design is sound. Table I shows the comparison of our model with two baseline LIVECell models. We found that our model slightly underperforms, albeit by a very small margin. This is not unexpected. Because the main goal of this study is to evaluate the CKS method, we have chosen model hyperparameters consistent with a light-weight model to facilitate experimentations. We demonstrate this point by comparing the inference speed of our model with the two others (Table I) and show that our model is roughly 6-8x faster.

TABLE I
PERFORMANCE OF SUPERVISED MODELS ON LIVECELL

| Models | Dice | AP50 | mAP | Speed (fps)* |
|---|---|---|---|---|
| Cellpose [2] | 0.831 | - | - | 1.85 |
| MaskRCNN [1] | 0.830 | 0.808 | 0.478 | 1.20 |
| Ours | 0.824 | 0.806 | 0.450 | 11.6 |

\* Inference speed measured with (520x704) images on Tesla V100.

In addition, we have performed ablation studies (Table II) to evaluate model design choices. Two key design choices of the principal model are to perform multi-scale cell detection and to employ a subnet to generate positional encodings. We thus evaluated models trained with single-scale cell detection (at s=4), and with constant encodings using a cosine encoding scheme (à la Transformer [48]). Our results indeed indicated that these design choices are important for the model's performance.

TABLE II
ABLATION STUDIES ON PRINCIPAL MODEL DESIGN

| Multi-Scale Detection | Position Encoding Net | Dice | AP50 | mAP |
|---|---|---|---|---|
| – | – | 0.793 | 0.780 | 0.423 |
| + | – | 0.805 | 0.788 | 0.437 |
| + | + | 0.824 | 0.806 | 0.450 |

### D. Point-supervised LIVECell Models

Fig. 4 demonstrates the training process of point-supervised models using the CKS methods. The model learned to predict cell locations fairly quickly. However, at this stage the principal model only recognized pixels immediately surrounding the detected location and produced inaccurate segmentations. Additionally, there is significant inconsistency between the principal model output and the collaborator model output, despite accessing the exact same input information. As the training continues, the consistency between the two models improves and so did the instance outputs from the principal model. The collaborator model only learned to predict cell-border at the very late stages of the training. But this was important for tuning the principal model output. Additionally, these examples showed that the accuracy of cell border predictions are relatively low.

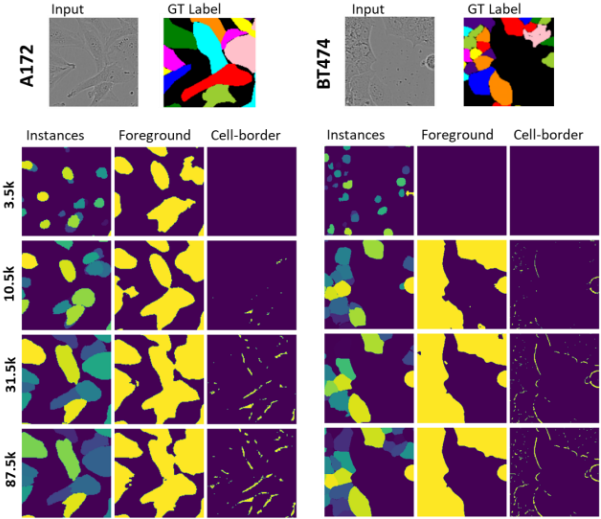

Fig. 4. CKS training progression. Top panel shows two example images. Bottom panel shows the model predictions at different training steps (left). The instance predictions are from the principal model and the foreground and cell-border are from the collaborator model.

Fig. 5 shows the final segmentation results by the CKS-trained models. Because different cell lines exhibit very different morphological characteristics, we provided one example for each cell line. Corresponding segmentations by the fully supervised model are also shown for comparison. These results qualitatively show that our method can establish single-cell segmentation models, despite using only the very weak point labels.

The quantitative evaluations of the model performances are shown in Table III. The results suggested that the point-supervised CKS models can achieve accuracy very close to that of a supervised model. However, the performances were uneven for different cell lines. Both the supervised model and CKS models struggle with non-convex shaped cells (e.g. SHSY5Y), indicating that the issue is likely related to the general approach of CNN, although the weak supervision exacerbates the difficulties. In addition, CKS models also have weakness with images with very low contrast (e.g. Huh7). Pre-training on Tissuenet dataset was clearly beneficial and for most cell lines the model obtained more than 90% of the performance of the supervised model, measured by AP50. However, for cells with relatively simple shapes (BV2 and SkBr3), pre-training led to very little differences.

In Table IV we compare our CKS model results with the two baseline methods on LIVECell dataset. Our results showed that the CKS models outperformed both baseline methods by a significant margin on all cell lines.

TABLE IV
COMPARISON OF POINT-SUPERVISED LIVECELL MODELS

| Dice | Nishimura-1 [47] | Nishimura-2 [47] | CKS-1 | CKS-2 |
|---|---|---|---|---|
| A172 | 0.624 | 0.678 | 0.761 | 0.772 |
| BT474 | 0.690 | 0.565 | 0.729 | 0.764 |
| BV2 | 0.541 | 0.643 | 0.806 | 0825 |
| Huh7 | 0.513 | 0.608 | 0.693 | 0.820 |
| MCF7 | 0.502 | 0.577 | 0.743 | 0.768 |
| SHSY5Y | 0.426 | 0.449 | 0.634 | 0.674 |
| SKOV3 | 0.612 | 0.649 | 0.779 | 0.848 |
| SkBr3 | 0.640 | 0.773 | 0.868 | 0.890 |
| ALL | 0.568 | 0.618 | 0.757 | 0.806 |

### E. Ablation Study

#### 1) Consistency loss function

Our training scheme employed an unorthodox linear loss function to enforce consistency between $M_p$ and $M_c$. In Table V we show comparison of our approach with the models trained instead with cross-entropy consistency loss (5). Our results showed that the cross-entropy loss resulted in significantly lower model accuracy as expected.

TABLE III
COMPARISON OF DIFFERENT TRAINING STRATEGIES

| | Dice | | | AP50 | | | AP75 | | | mAP | | |
|---|---|---|---|---|---|---|---|---|---|---|---|---|
| | CKS-1* | CKS-2* | Supervised | CKS-1* | CKS-2* | Supervised | CKS-1* | CKS-2* | Supervised | CKS-1* | CKS-2* | Supervised |
| A172 | **0.761** | **0.772** | 0.802 | 0.687 | **0.721** | 0.764 | 0.171 | 0.227 | 0.349 | 0.277 | 0.312 | 0.381 |
| BT474 | **0.729** | **0.764** | 0.782 | 0.688 | **0.713** | 0.769 | 0.247 | 0.319 | 0.438 | 0.304 | 0.351 | 0.430 |
| BV2 | **0.806** | **0825** | 0.830 | **0.878** | **0.889** | 0.900 | 0.534 | 0.563 | 0.639 | 0.495 | **0.510** | 0.552 |
| Huh7 | **0.693** | **0.820** | 0.831 | 0.448 | 0.514 | 0.781 | 0.031 | 0.108 | 0.515 | 0.100 | 0.188 | 0.473 |
| MCF7 | **0.743** | **0.768** | 0.778 | **0.744** | **0.753** | 0.789 | 0.193 | 0.252 | 0.356 | 0.308 | 0.333 | 0.392 |
| SHSY5Y | **0.634** | **0.674** | 0.703 | 0.313 | 0.367 | 0.548 | 0.004 | 0.008 | 0.100 | 0.072 | 0.087 | 0.199 |
| SKOV3 | **0.779** | **0.848** | 0.864 | 0.740 | **0.804** | 0.876 | 0.159 | 0.305 | 0.554 | 0.294 | 0.376 | 0.509 |
| SkBr3 | **0.868** | **0.890** | 0.898 | **0.934** | **0.938** | 0.951 | **0.749** | 0.717 | 0.821 | **0.617** | **0.604** | 0.667 |
| ALL | **0.757** | **0.806** | 0.824 | 0.712 | **0.734** | 0.806 | 0.304 | 0.333 | 0.468 | 0.342 | 0.361 | 0.450 |

* CKS-1 and CKS-2 denote weakly-supervised models under initialization scheme 1 and 2 respectively.
Rows are different cell lines. Bold font denotes metrics that reached >90% of the performance of the supervised model.

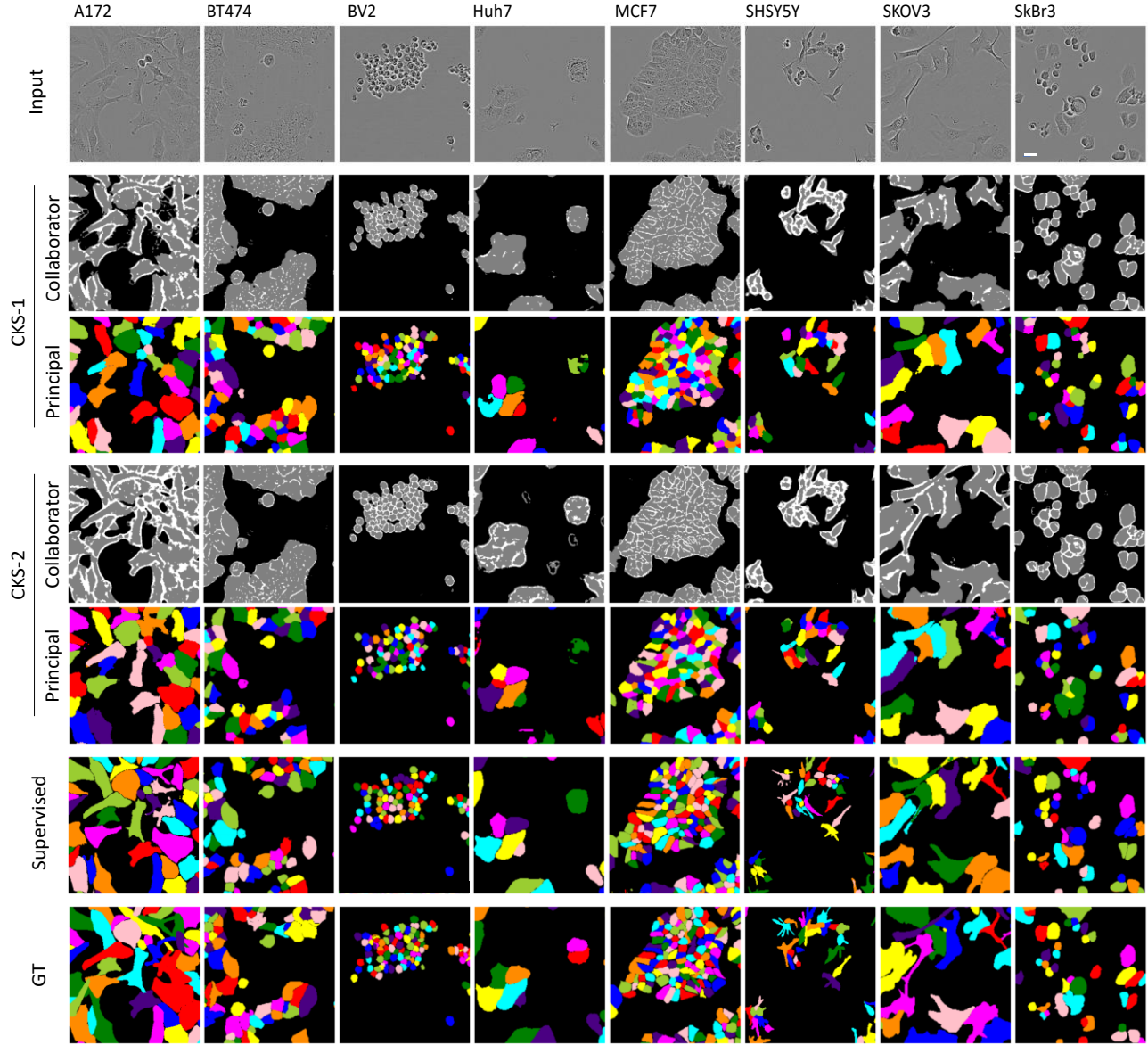

Fig. 5. Examples of trained model predictions on LIVECell data. For point-supervised models (CKS-1 and CKS-2) the predictions from the collaborator models, including the cell foreground(gray) and cell border (white) are shown together with principal model predictions (colored).

TABLE V
ABLATION STUDIES ON CONSISTENCY LOSS

| Principal Model | Collaborator Model | Dice | AP50 | mAP |
|---|---|---|---|---|
| Entropy | Linear | 0.655 | 0.530 | 0.227 |
| Linear | Entropy | 0.698 | 0.610 | 0.281 |
| Linear | Linear | 0.806 | 0.734 | 0.361 |

### 2) Sensitivity to hyperparameter $d$

Point label has an inherent problem: Considering a case in which a cell organelle (e.g., nucleus) is also generally located at cell center, there is no inherent reason for the model to produce cell segmentation instead of organelle segmentation, as they would have been labeled the same. We got around this problem by using $d$ to specify prior knowledge regarding the expected cell sizes. Here we test how sensitive the model results depend on $d$. Beside the original model, we tested two additional conditions: (1) we set the same $d$ values (20) for all cell lines, and (2) we remove the prior component completely by setting $\pi = 0$. Table VI presents the results. The main finding here is that the exact choices of $d$ values have a minor impact on model accuracy, but removing the prior specification completely resulted in significant decrease in model accuracy.

TABLE VI
IMPACT OF HYPERPARAMETER $d$

| | AP50 | AP75 | mAP |
|---|---|---|---|
| NOT USED | 0.482 | 0.211 | 0.225 |
| CONSTANT $d$ | 0.714 | 0.238 | 0.316 |
| PER-CELLLINE $d$ | 0.734 | 0.333 | 0.361 |

### 3) Sensitivity to collaborator model capacity

One distinct feature of CKS is that the consistency is evaluated between two highly unbalanced models. The collaborator model is designed to be very light weight. Table VII presents comparisons of some other collaborator model designs with various degrees of complexity (Fig. 6). Specifically, we tested deeper U-Net for image foreground prediction and more complex CNN for cell border prediction. The results shown here demonstrated that in general, increasing the collaborator complexity did not help with improving the principal model accuracy. In fact, increasing the model complexity for cell border detection had a clear adverse effect. These are characteristics generally unexpected for collaborative learnings [43], suggesting that the mechanism of CKS may be different from other related methods.

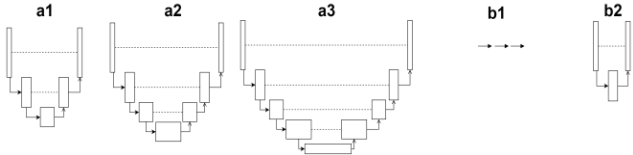

Fig. 6 Schematics of network structures used for collaborator model. Net a1-a3 were used for predicting foreground pixels. Net b1 and b2 were used for cell border pixel. The b1 is a simple three-layer CNN and the rest are U-Net. Every U-Net structure starts with a 16-channel layer and doubles the channel width on every down scaling operation.

TABLE VII
IMPACT OF COLLABORATOR MODEL COMPLEXITY

| Foreground Pixel | Cell Border | AP50 | AP75 | mAP |
|---|---|---|---|---|
| a1 | b1 | 0.734 | 0.333 | 0.361 |
| a2 | b1 | 0.734 | 0.325 | 0.358 |
| a3 | b1 | 0.733 | 0.317 | 0.346 |
| a1 | b2 | 0.706 | 0.186 | 0.294 |
| a3 | b2 | 0.593 | 0.137 | 0.249 |

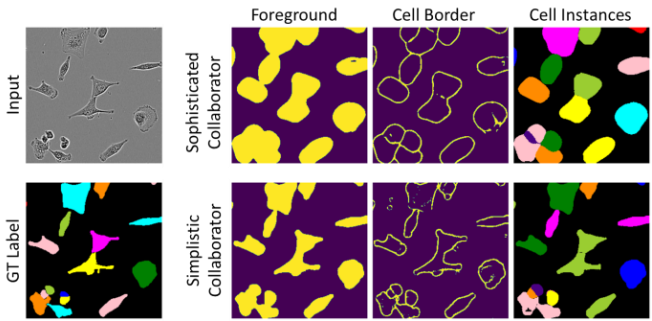

Fig. 7 Comparison of "sophisticated" collaborator setup with "simplistic" collaborator setup. Cell instance predictions are from the principal models and foreground and cell border are from the collaborator models.

To further elucidate why a more complex collaborator model would result in lower overall performance, we manually examined the individual model predictions. Fig. 7 shows an example comparing a "sophisticated" collaborator setup (a3/b2 configuration) with the "simplistic" collaborator setup (a1/b1). It is evident that in both cases, the model training reaches good "consistency" between the principal model and the collaborator model outputs. However, in the case of "sophisticated" collaborator, the "consistency" does not translate to "correctness". Instead, both the principal model and the collaborator model are "consistently" wrong. This observation agrees with the notion that the collaborator model should be viewed as a regularization function. Its main purpose is to regularize/restrict the search space of the principal model. In this context, a very sophisticated collaborator has a lesser regularization power due to its own degrees of freedom.

### F. Extending to Nuclei Image Derived Point Labels

For all experiments in LIVECell dataset, we generated point label from centroids of segmentation masks. It is unclear whether the model took advantage of this hidden correlation. Therefore, we also test our method on the A431 dataset, for which the point labels were generated using an unsupervised blob detection algorithm based on nuclei counter-statin data, and thus not at the exact center of the cells. In addition, this training scheme can be considered as completely unsupervised, which has practical advantages.

We presented examples of segmentation results in Fig. 8 and quantitative evaluations in Table VIII. These results showed that CKS method can indeed be used for "inexact" point labels, such as those indicated by the nuclei locations. In addition, CKS models outperformed the baselines methods in all metrics evaluated here.

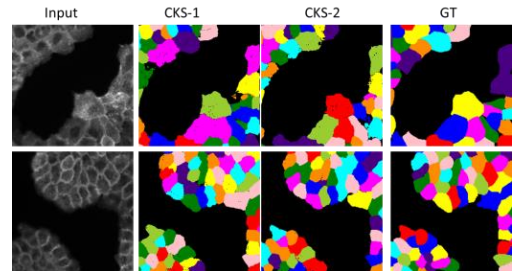

Fig. 8 Examples of cell segmentation results on A431 dataset.

TABLE VIII
MODEL PERFORMANCE ON A431 DATASET

|  | Dice | AP50 | AP75 | mAP |
|---|---|---|---|---|
| Nishimura-1 | 0741 | 0.571 | 0.025 | 0.167 |
| PDAM | 0.702 | 0.384 | 0.044 | 0.125 |
| CKS-1 | 0.821 | 0.887 | 0.301 | 0.407 |
| CKS-2 | 0.823 | 0.897 | 0.317 | 0.418 |

Among the two baseline methods, the weakness of the PDAM results is a surprise. Upon further investigation, we found that the PDAM method had struggled to achieve good results even on the source domain data. We attributed the cause to the idiosyncratic labeling of the TissueNet dataset. As shown in Fig. 9, the original TissueNet data comprised of two-channel microscopy images of membrane staining and nucleus staining. While most of the labeling outlines the whole cell, in areas with low membrane signal, the labeler chose to outline the nucleuses instead. The PDAM method uses CycGAN [49] to adapt source domain images to resemble the target domain, which are single-channel images of membrane/cytoplasmic staining. This appears to cause confusion to the segmentation model as it is now forced to perform segmentation in areas with little to no contrast. This experiment exposed a weakness in UDA type methods, as they have a dependence on the nature of both source domain and target domain and can lead to unexpected complications. In contrast CKS is a straightforward gradient descent-based training pipeline that cares only about the target domain data.

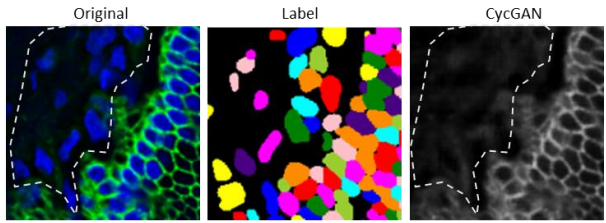

Fig. 9 Idiosyncratic labeling causes difficulties in domain adaptation training. Left and middle panels show an example of source domain image and the corresponding labeling. The right panel shows the CycGAN adaptation of the original image. The dotted line outlines the region of extremely weak contrast after adaptation.

## V. CONCLUSIONS AND DISCUSSIONS

In conclusion, we proposed in this paper a novel method to train single-cell segmentation models using only point labels. The proposed method is, to our knowledge, the first one of this type that performs segmentation in an end-to-end fashion. Below we briefly discuss the pro and cons of the proposed method:

Pros:
1. The collaborator model is very light weight and adds very little in terms of additional computational complexity, compared to training the principal model alone.
2. Model performance can be very close to fully supervised ones if the shape complexity of the cell is low.

Cons:
1. The method relies on a fixed hyperparameter to decide how big of an area to perform the segmentation for, which is not computationally efficient if the inputs data contains both very large and very small cells, because the hyperparameter needs to be set according to the largest cells.
2. The model performance is uneven and is dependent on the cell shape complexity. It is possible that this problem can be alleviated in semi-supervised training settings where part of the data is labeled with segmentations.

A final point worth noting is that training a model to predict point label is a powerful surrogate task that allows the model to learn useful instance-level features. Indeed, for a model to be able to predict cell locations, it must have internal knowledge about what a "cell" is already. What is missing is an algorithm to map the abstract image features to a tangible output that is segmentation. From this point of view, the CAM-based methods (e.g., Nishimura-1) are essentially rational, hand-crafted algorithms to achieve this mapping, and CKS is the machine-learning counterpart. Since the assumption is that the encoder already produces good image features, the major task here is therefore not about learning more image features, but to constrain the segmentation head to produce a desired output. This is probably the main reason why a regularization scheme, such as CKS, worked well here.